\pdfoutput=1
\documentclass{article}
\usepackage{spconf,amsmath,graphicx,hyperref}
\hypersetup{hidelinks}
\usepackage{booktabs}
\usepackage{caption}
\usepackage{expex}
\usepackage{tipa}
\usepackage{multirow}

\usepackage{fancyhdr}
\fancypagestyle{FirstPage}{
}

\usepackage{xcolor}

\title{FRUSTRATINGLY EASY DATA AUGMENTATION FOR LOW-RESOURCE ASR}

\twoauthors
 {Katsumi Ibaraki\thanks{Our code is available at \url{https://github.com/kibaraki/frustratingly-easy-asr-augmentation}}}
	{University of Notre Dame, USA \\
	Dept.~of Computer Science and Engineering\\
	\texttt{kibaraki@nd.edu}}
 {David Chiang}
	{University of Notre Dame, USA \\
	Dept.~of Computer Science and Engineering\\
	\texttt{dchiang@nd.edu}}
\begin{document}

\ninept
\maketitle

\begin{abstract}
This paper introduces three self-contained data augmentation methods for low-resource Automatic Speech Recognition (ASR). 
Our techniques first generate novel text---using gloss-based replacement, random replacement, or an LLM-based approach---and then apply Text-to-Speech (TTS) to produce synthetic audio. 
We apply these methods, which leverage only the original annotated data, to four languages with extremely limited resources (Vatlongos, Nashta, Shinekhen Buryat, and Kakabe). 
Fine-tuning a pre-trained Wav2Vec2-XLSR-53 model on a combination of the original audio and generated synthetic data yields significant performance gains, including a 14.3\% absolute WER reduction for Nashta. 
The methods prove effective across all four low-resource languages and also show utility for high-resource languages like English, demonstrating their broad applicability.

\end{abstract}
\section{Introduction}
\thispagestyle{FirstPage}
\label{sec:intro}
Despite substantial improvements in Automatic Speech Recognition (ASR), state-of-the-art models remain dependent on large-scale datasets; most progress has centered around high-resource languages \cite{10.5555/3495724.3496768}.
Even research in low-resource settings often uses additional text data, such as dictionaries, for data processing or synthesis. 
However, these resources may be unavailable for endangered or less-studied languages, and new data collection is often infeasible due to a scarcity of speakers, the remote geographic locations where they reside, or other logistical barriers.
This paper targets this gap by introducing simple data augmentation methods that expand a corpus using only its existing annotated data.
We propose three techniques:
(1) replacing words by words with the same gloss, (2) replacing words with random words, and (3) generating new sentences using a Large Language Model (LLM).
We use these methods to synthesize new sentences, which are then converted into synthetic speech audio using a Text-to-Speech (TTS) model.
Our replacement methods are an extension of semantic- and template-based replacement techniques \cite{wei-zou-2019-eda, hou-etal-2018-sequence}, adapted to work without external lexical databases, and allowing simultaneous, multi-word substitutions.
We evaluate our methods on four low-resource languages---Vatlongos, Nashta, Shinekhen Buryat, and Kakabe---and on the high-resource English LibriSpeech \cite{panayotov2015librispeech}, to assess their broader applicability.

In summary, we make three contributions.
First, we present an approach that combines text and audio data augmentation to generate synthetic training data from a labeled corpus.
Second, we develop ASR models for Vatlongos, Nashta, Shinekhen Buryat, and Kakabe, which, to our knowledge, are the first ASR systems for each of these languages. 
Third, we demonstrate that for ASR in low-resource settings, prioritizing phonemic and structural variety can be more effective for model training than preserving semantic coherence.

\begin{figure*}
    \centering
    \includegraphics[width=1.0\linewidth]{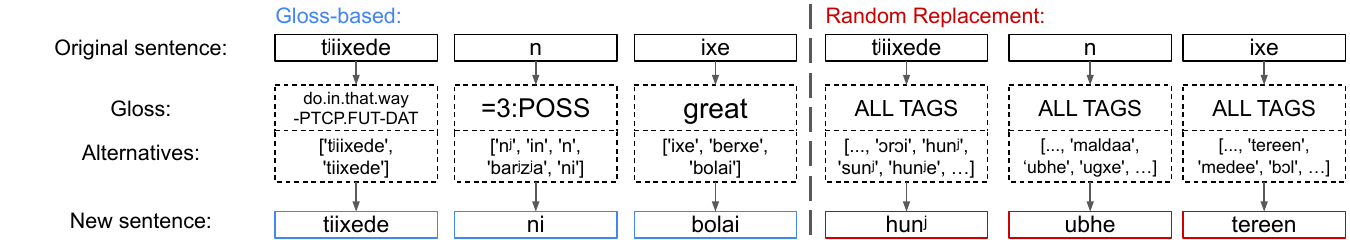}
    \caption{
    Illustration of gloss-based replacement (left) and random replacement (right) applied to the beginning of the Shinekhen Buryat sentence, ``Then the first queen told him.'' 
    The gloss-based method replaces each word with an alternative from the set of all words sharing the same gloss in the training data.
    In contrast, the random replacement method 
    ignores all linguistic information, substituting
    each word with a random selection from all words in the training data.
    }
    \label{fig:gloss-based}
\end{figure*}

\section{Related Work}
Data augmentation is a well-established technique for improving text-based models.
Common strategies range from simple operations---such as synonym replacement via WordNet \cite{miller_wordnet} and random word insertion, swapping, or deletion \cite{wei-zou-2019-eda}---to more sophisticated methods.
These include using bidirectional LSTMs for contextual augmentation \cite{kobayashi-2018-contextual} and round-trip translation to paraphrase text for machine translation \cite{xie2020unsupervised}.

In ASR, these text augmentation techniques are combined with speech synthesis to expand the training corpora.
For instance, Zevallos et al. \cite{zevallos22_interspeech}, using delexicalization methods \cite{hou-etal-2018-sequence}, substitute words within predefined semantic frames.
While effective, the applicability of these methods is limited.
First, the substitutions are typically restricted to a few common semantic categories within a sentence, which makes most of the sentence structure unchanged.
Second, and more critically for our purposes, these techniques are dependent on external lexical or semantic resources to identify and supply replacements.
More recently, LLMs have been used for synthetic data generation.
Previous work has used LLMs for relatively constrained tasks, such as paraphrasing text to improve $n$-gram language models \cite{Nagano}, or generating in-domain data for adaptation to new domains \cite{Su2024}.
While more unconstrained, novel sentence generation has been explored in other contexts, its application specifically as a data augmentation technique for low-resource ASR, especially for languages that are truly unseen or typologically distant from a model's pre-training data, has received less attention.

\section{Method}

\subsection{Synthetic Text Generation}
\label{subsec: synthetic text generation}
We explore three different methods for synthetic text generation: gloss-based replacement, random replacement, and LLM-based generation.
The two replacement methods are shown in Figure \ref{fig:gloss-based}.

\textbf{Gloss-based:} 
Previous work \cite{zevallos22_interspeech, hou-etal-2018-sequence} has typically utilized datasets with consistent sentence structures and high semantic overlap, allowing for the extraction of frames like \textit{distance} or \textit{location}. 
In contrast, our data consists of more unstructured, freely spoken speech. 
This lack of regularity makes template-based augmentation impractical without extensive, handcrafted rules, and the varied contexts make it ineffective to focus only on frequent tags. 
Therefore, we adopt a more comprehensive strategy: we treat every word position as a candidate for substitution and use glosses (brief definitions and/or grammatical tags) instead of formal semantic frames, which would require nonexistent external resources.
A dictionary is constructed from the training data by mapping each gloss to a list of all words that share it, based on the provided gloss and POS information.
Treating each original sentence as a template for guiding 
gloss order for each sentence, we refer to the dictionary to randomly select a possible alternative, and continue the process for each position.
Although the sentences maintain the same glosses as the original, this process does not ensure that the sentences are grammatically correct or are natural-sounding to a native speaker.
While we acknowledge the limitations of this method, given the limited resources for these languages, this approximation is a pragmatic approach.

\textbf{Random Replacement:} 
As a more extreme variant of gloss-based replacement, we also introduce a random replacement method. 
In this approach, each original sentence serves as a structural template solely to define the length of the new sentence. 
For each word position, a replacement is randomly sampled from the entire vocabulary of the training data, completely disregarding any gloss or semantic information. 
While this technique ensures that all generated words are in-vocabulary, the resulting sentences are not expected to be grammatical or semantically coherent.

\begin{figure}
    \centering
    \fbox{
        \begin{minipage}{0.46\textwidth}
        \ttfamily\footnotesize
        Given the following CSV, focus on columns [text, clean\_text, english, gloss] and generate \textbf{\{number of sentences in train\}} sentences in a CSV with all of the original columns, consisting of only the new sentences; this is in \textbf{\{language\}}, \textbf{\{language description\}}; do not use Python code to generate the sentences but rather use your understanding of other languages as an LLM to generate sentences; make sure that the text and gloss generated match; this text will be passed on to a TTS model to generate synthetic audio, to use for additional training data for a wav2vec2-based ASR model.
        \end{minipage}
    }
    \caption{LLM prompt for generating synthetic sentences. 
    }
    \label{fig:simple-prompt}
\end{figure}

\textbf{LLM-based:} 
In addition to replacement-based techniques, we use Google's Gemini 2.5 Pro \cite{Google2025_Gemini} to generate entirely new sentences. 
Our approach deliberately leverages the model's capacity for hallucination, a typically avoided behavior, to create novel words and syntactic structures not present in the original small training set. 
The model was prompted with the existing training data, including transcriptions, translations, and gloss/POS information, as detailed in Figure \ref{fig:simple-prompt}. 
This generative method was pursued after determining that other models, such as Open\-AI's ChatGPT, were unsuitable, as they tended to produce code for rigid, template-based sentences using limited semantic categories, rather than the desired novel text.

\subsection{Synthetic Audio Generation}
\label{subsec: synthetic audio generation}

The three types of synthetic text were converted into speech using Kokoro, an open-weight TTS model that can process both IPA and orthographic inputs.\footnote{\url{https://huggingface.co/hexgrad/Kokoro-82M}} 
For this synthesis, we used a standardized set of five voices across all languages rather than attempting to clone the voices of the original speakers. 
This decision was based on two factors. 
First, the limited audio per speaker was insufficient for fine-tuning individual, high-quality TTS models; using a uniform set of voices ensures experimental consistency. 
Second, preliminary experiments in voice conversion, where each original speaker's voice was used to generate all training sentences, did not yield any performance improvements.

\begin{table*}
\centering\small
\begin{tabular}{@{}llllllllll@{}}
\toprule
 & Minutes & Speakers & Total Words & Total Unique & Train Words & Train Unique & Gloss & \% Alt. & \% Out\\
\midrule
 Vatlongos & 286  & 60 &66984 &3644 & 23315 &2823  & 422 &  10.0 & 55.4 \\

 Nashta  & 48  & 4 & 5473 & 1643 & 3982& 886  & 637 & 29.5 & 95.3\\

  Kakabe & 99  & 34 & 17798 & 1830 & 12512& 1525 & 259 & 24.3 & 20.5 \\

 Shinekhen Buryat & 97  & 4 & 11338 & 3089 &7895& 2276 & 2242 &  11.4 & 60.0\\

 \midrule
 English \\
 \quad LibriSpeech-54 
 & 54 & 40& 8738& 2575& 6164 & 2039 & 396 & 55.8 & 99.8\\
 \quad LibriSpeech-108 
 & 108 &40 & 17449& 4212& 12236 & 3326 & 491 & 58.9 & 99.9\\
 \quad LibriSpeech-324 
 & 324 & 40 & 52576 & 8138 & 36842 & 6644 & 592 & 63.9 & 99.9 \\
  \quad LibriSpeech-1207 
  & 1207 & 251 & 198291 & 17243 & 137411 & 14263 & 902 & 67.8 & 99.9\\
\bottomrule
\end{tabular}
\caption{
Summary of corpus statistics for each language.
The table shows the number of minutes of transcribed audio, speakers, total and unique words in the corpus, total and unique words in training, gloss tags in training, percentage of glosses with substitution alternatives in training, and the out-of-vocabulary rate for LLM-generated text relative to the training split. 
The LibriSpeech splits are marked with the number of minutes; the first three are from \texttt{test-clean} and 1207 is from \texttt{train-clean-100} as labeled in \cite{panayotov2015librispeech}.}
\label{tab:metadata}
\end{table*}


\subsection{Fine-tuning Wav2Vec2}
\label{subsec: finetuning wav2vec2}
We fine-tune the pre-trained Wav2Vec2-XLSR-53 model \cite{conneau2020unsupervised} using a Connectionist Temporal Classification (CTC) loss \cite{10.1145/1143844.1143891}. 
Hyperparameters were selected via preliminary experiments to prevent overfitting: we used a learning rate of $1.0 \times 10^{-4}$ and set language-specific training epochs ranging from 10 to 50. 
To ensure a controlled experiment, we excluded other augmentations like SpecAugment \cite{park19e_interspeech} and maintained a 1:1 ratio of synthetic to original data, as initial tests showed larger proportions of synthetic data harmed performance.

\section{Experiment Setup and Results}

\subsection{Data}
We focus on four low-resource languages and dialects, all with five hours or less of audio data.
To our knowledge, all transcriptions were made manually, without the use of grapheme-to-phoneme tools.
All audio clips were downsampled to 16 kHz for training.

For the data splits, we allowed speaker overlap, following Wei et al. \cite{WEI20221} and Liu et al. \cite{liu-etal-2023-investigating}, who demonstrated that this has a negligible impact in low-resource contexts. 
While the level of transcription detail varies across our data sources, we treat all IPA-based transcriptions as phonemic representations and refer to the units as \textit{phonemes}. 
This aligns with the objective of our ASR models, which is to learn the contrastive sound system of each language.

\textbf{Vatlongos} is a language of Vanuatu, an island nation in the South Pacific, with an estimated 3,000 speakers \cite{ridge2018language, Crowley01042000}. 
For this study, we use a corpus collected by Ridge (Massey University) and hosted on the Pangloss Collection.\footnote{\url{https://pangloss.cnrs.fr/corpus/Vatlongos?lang=en}} 
Each utterance is annotated with a text transcription (orthography and broad IPA conversion), English and Bislama translations, and gloss. 
An example is shown below:

\pex[aboveglftskip=0ex]<withparts> 
\begingl 
\gla Dilamun ba biteni mama nan//
\glb 3s.return.ind 3s.go.ind 3s.say.ind mother cl.gen-3s.poss//
\glft `He turned back and told his mother.'//
\endgl
\xe

\textbf{Nashta} is an under-documented South Slavic variety spoken in the Balkans that is related to literary Bulgarian and Macedonian and shows Greek influence \cite{adamou2022nashta}. 
The data comes from a corpus on the Pangloss Collection,\footnote{\url{https://pangloss.cnrs.fr/corpus/Nashta?lang=en}} collected by Adamou (National Centre for Scientific Research in France), who established ``Nashta'' as a scholarly convention for this variety.
The dataset 
is annotated with broad IPA transcriptions, French and English translations, and gloss. 
The following example is taken from this corpus:

\pex[aboveglftskip=0ex]<withparts> 
\begingl 
\gla i tam \textipa{"}setne ka\textipa{"}va \textipa{"}xodexa tam \textipa{"}nare//
\glb and there then when go.IPFV there up //
\glft `And then, when they would go to the mountain'//
\endgl
\xe


\textbf{Shinekhen Buryat} is a Mongolic language variety spoken by approximately 6,000 people in Inner Mongolia, China \cite{yamakoshi2011three}. 
It is a variant of the broader Buryat language, which has around 520,000 speakers across Russia, China, and Mongolia \cite{TheShinekhenBuryatVarietyPreliminaryAnalysis}. 
The data for this study was collected and transcribed by Yamakoshi (Tokyo University of Foreign Studies).\footnote{ 
\url{https://tufs.repo.nii.ac.jp/search?search_type=2&q=1729497608274}
} 
The corpus 
is annotated with a more narrow IPA transcription, English and Japanese translations, and gloss. 
An example from the corpus is provided below:

\pex[aboveglftskip=0.5ex,extraglskip=0ex]<withparts> 
\begingl 
\gla ii-g-eed        zalg-aad        {\textopeno}d{\textopeno}{\textopeno}    ajan    x\textipa{8}\textipa{8}-z\textsuperscript{j}e//
\glb do.like.this-E-CVB.PFV        continue-CVB.PFV        now     journey:INDF    follow-CVB.IPFV//
\glft `Then he continued his journey.'//
\endgl
\xe

\textbf{Kakabe} is spoken in Guinea, with approximately 50,000 speakers, and belongs to the Mande languages, but has influence from Pular \cite{vydrina2017corpus}. 
We use the corpus collected by Vydrina (National Centre for Scientific Research in France), hosted on the Pangloss Collection.\footnote{\url{https://pangloss.cnrs.fr/corpus/Kakabe?lang=en}}
The corpus contains a broad IPA transcription, English translations, and gloss.
An example from the corpus is shown below:

\pex[aboveglftskip=0ex]<withparts> 
\begingl 
\gla a \textipa{f\'\textopeno\textopeno} \textipa{k\'aa,} \textipa{\'oma} {n'} a \textipa{f\'\textopeno\textopeno} \textipa{k\'elen} \textipa{k\'elen}//
\glb 3SG say this.way 1PL.INCL OPT 3SG say one one //
\glft `Tell that we have to speak one by one.'//
\endgl
\xe


\textbf{LibriSpeech} is a corpus of read English speech
\cite{panayotov2015librispeech}.
To evaluate our methods on a high-resource language which the model has seen during pre-training, we use the \texttt{test-clean} and \texttt{train-clean-100} splits from this corpus.
To emulate the documentation-style corpora, we use \texttt{spaCy} to add POS and dependency tags.

\begin{table*}[t]
\centering
\fontsize{9}{11}\selectfont 
\setlength{\tabcolsep}{4pt} 

\begin{tabular}{@{}ll *{4}{cccc}@{}}
\toprule
& & \multicolumn{4}{c}{Vatlongos} & \multicolumn{4}{c}{Nashta} & \multicolumn{4}{c}{Shinekhen Buryat} & \multicolumn{4}{c}{Kakabe} \\
\cmidrule(lr){3-6} \cmidrule(lr){7-10} \cmidrule(lr){11-14} \cmidrule(lr){15-18}
& & Base & Gloss & Rand. & LLM & Base & Gloss & Rand. & LLM & Base & Gloss & Rand. & LLM & Base & Gloss & Rand. & LLM \\
\midrule
\multirow{2}{*}{PER/CER} & Val & 16.0 & 14.8 & \textbf{14.4 } & 15.2 & 26.4 & 22.3* & \textbf{22.0}* & 22.5* & 13.9 & 13.2 & \textbf{13.0}* & 13.4 & 23.7 & 23.2 & \textbf{23.0 } & {23.4 } \\
& Test & 16.3 & 15.3* & \textbf{14.9}* & 15.3 & 24.9 & \textbf{20.0}* & \textbf{20.0}* & 20.3* & 14.9 & \textbf{14.3} & {14.9 } & 15.0 & 22.4 & 22.0 & \textbf{21.4}* & 21.6* \\
\midrule
\multirow{2}{*}{WER} & Val & 48.8 & 44.1* & \textbf{42.7}* & 45.1* & 77.2 & 65.3* & \textbf{64.7}* & 64.8* & 46.1 & \textbf{44.4} & 44.7 & 45.9 & 52.6 & \textbf{51.1} & {51.9 } & {53.7 } \\
& Test & 47.4 & \textbf{43.0}* & \textbf{43.0}* & 43.7* & 75.4 & \textbf{61.1}* & 63.2* & 66.1* & 48.1 & \textbf{44.6} & 44.9 & 47.7 & 50.5 & 49.7 & 49.3* & \textbf{49.1}* \\
\bottomrule
\end{tabular}

\vspace{3ex} 

\begin{tabular}{@{}ll *{4}{cccc}@{}}
\toprule
& & \multicolumn{16}{c}{English} \\
& & \multicolumn{4}{c}{LibriSpeech-54} & \multicolumn{4}{c}{LibriSpeech-108 } & \multicolumn{4}{c}{LibriSpeech-324} & \multicolumn{4}{c}{LibriSpeech-1207} \\
\cmidrule(lr){3-6} \cmidrule(lr){7-10} \cmidrule(lr){11-14} \cmidrule(lr){15-18}
& & Base & Gloss & Rand. & LLM & Base & Gloss & Rand. & LLM & Base & Gloss & Rand. & LLM & Base & Gloss & Rand. & LLM \\
\midrule
\multirow{2}{*}{PER/CER} & Val & 6.6 & 8.7 & 8.2 & \textbf{4.8}* & 4.6 & 5.8 & 5.0 & \textbf{3.5 } & 4.3 & \textbf{2.7}* & 4.0* & 2.8* & 1.9 & 1.6 & \textbf{1.5 } & 1.9 \\
& Test & 6.7 & 8.7 & 8.2 & \textbf{4.6}* & 4.5 & 5.7 & 4.9 & \textbf{3.6}* & 4.2 & 2.9* & {4.0 } & \textbf{2.7}* & 1.8 & 1.6 & \textbf{1.4}* & 1.8 \\
\midrule
\multirow{2}{*}{WER} & Val & 24.9 & 32.7 & 30.9 & \textbf{17.2}* & 17.4 & 22.3 & 18.8 & \textbf{12.4}* & 16.7 & \textbf{10.0}* & 15.4* & \textbf{10.0}* & 6.9 & 5.6 & \textbf{5.3}* & 6.6 \\
& Test & 24.9 & 31.9 & 30.6 & \textbf{16.3}* & 16.7 & 21.5 & 18.4 & \textbf{12.6}* & 16.4 & 10.5* & 15.4* & \textbf{10.3}* & 6.7 & 5.5 & \textbf{5.1}* & 6.3 \\
\bottomrule
\end{tabular}

\caption{
The results for all models.
All values in the table are in percentage (\%). 
The improvements that are statistically significant from the baseline (at significance level of 0.05), using paired bootstrap resampling \cite{bisani_bootstrap, koehn-2004-statistical}, are indicated with an asterisk (*).
}
\label{tab:1-flipped-combined}
\end{table*}

\subsection{Evaluation}

We evaluate model performance using three metrics: Word Error Rate (WER), Character Error Rate (CER), and Phoneme Error Rate (PER). 
CER is computed for our orthography-based models, while PER, which is analogous to CER but operates on phonemes, is computed for our IPA-based models. 
For cross-linguistic analysis, both CER and PER are reported, though we acknowledge these metrics are not directly comparable.
Additionally, we assess the statistical significance of improvements over the baseline for all metrics using paired bootstrap resampling \cite{bisani_bootstrap, koehn-2004-statistical}, with 10,000 samples and a significance level of 0.05.

\subsection{Results}
The results are displayed in Table \ref{tab:1-flipped-combined}. The improvements of the metrics mentioned below are based on the test set results.

For Vatlongos, all three data augmentation techniques improved performance over the baseline across all metrics, and we observed no differences between using IPA or orthography for sentence generation.
The most effective method was random replacement, which achieved a 1.4\% absolute (8.8\% relative) reduction in PER and a 4.4\% absolute (9.4\% relative) reduction in WER. 
The gloss-based and LLM-based methods also yielded strong results, with the former achieving a WER reduction of 4.4\% (9.4\% relative). 
Across all methods, the relative improvement in WER was larger than in PER. 

With Nashta, all methods yielded significant improvements, with gloss-based and random replacement performing best. 
The gloss-based method reduced PER by 4.9\% (19.8\% relative) and WER by 14.3\% (18.9\% relative). 
Notably, and in contrast to other languages, the relative improvements in PER for Nashta were comparable to those in WER, suggesting the augmentations provided balanced benefits at both the phoneme and word levels.

%
%
\label{sec:shibur-result}

Of the low-resource languages tested, Shinekhen Buryat produced the most mixed results, with the effectiveness of the augmentation varying significantly by method. 
The gloss-based approach performed best, improving PER by 0.6\% (4.1\% relative) and WER by 3.5\% (7.3\% relative) over the baseline. 
Random replacement also provided an improvement; in contrast, the LLM-based method had a much smaller effect when compared to the other languages.
For the two replacement methods, the relative gains in WER were more notable than in PER.

For Kakabe, we had statistically significant, though more modest, improvements.
Random replacement was the most effective, achieving a PER reduction of 1.0\% (4.6\% relative) and a WER reduction of 1.2\% (2.3\% relative). 
The gloss-based replacement and LLM-based approach yielded similar results.

For LibriSpeech, the LLM-based augmentation provided consistent improvements,
while the replacement methods underperformed at smaller data subsets.
However, on larger subsets, both replacement methods became highly effective. 
Gloss-based replacement achieved a reduction of up to 1.3\% (31.5\% relative) in PER and 5.9\% (36.1\% relative) in WER for LibriSpeech-324, and random replacement achieved a reduction of up to 0.4\% (22.5\% relative) in PER and 1.5\% (22.6\% relative) in WER for LibriSpeech-1207.

\section{Discussion}

The effectiveness of the three proposed augmentation methods varied significantly across different data resource conditions.
In the four low-resource languages, the simple, self-contained replacement methods were the most reliable source of performance gains. 
The choice between the two methods appears language-dependent, with the random method excelling for Vatlongos, Nashta, and Kakabe, and the gloss-based method for Shinekhen Buryat.
Contrary to expectations from prior work emphasizing semantic consistency, the fully random replacement method was marginally superior to the more structured gloss-based approach for several tests. 
This leads to a takeaway: in extremely data-scarce ASR settings, increasing the sheer variation of phonemic and structural patterns---even when constrained to the existing vocabulary---can be more beneficial for model training than preserving semantic coherence. 
In the high-resource LibriSpeech setting, both replacement methods were initially detrimental, only becoming beneficial after the dataset size was larger than any of the low-resource languages' datasets. 

The LLM-based approach yielded more varied outcomes. 
It performed competitively for some languages, while underperforming the baseline or not improving for others. 
We speculate that its unique strength lies in its ability to hallucinate novel words not present in the training corpus, which may help the model generalize to unseen vocabulary. 
However, its unreliability makes it a higher-risk strategy. 
With LibriSpeech, it was the most consistent across different data sizes, but we note that because it was in English, the LLM did not need to synthesize in the same way as the other languages.
Ultimately, simple and easily replicable methods like gloss-based and random replacement offer a reliable way to expand datasets and improve low-resource ASR. 

\section{Conclusion}

This study evaluated three data augmentation techniques---gloss-based replacement, random replacement, and LLM-based gen\-eration---for low-resource ASR by generating synthetic data for four languages. 
All methods yielded significant performance gains, most notably for Nashta, where our best approach achieved absolute reductions of 4.9\% in PER and 14.3\% in WER.
Meaningful WER reductions were also achieved for 
the other three languages.
Furthermore, we demonstrated the versatility of these techniques by successfully applying them to English, confirming they are viable strategies for improving Wav2Vec2-based ASR models in data-scarce environments.


\section{Acknowledgements}
The authors have no relevant financial or nonfinancial interests to disclose.

\section{Compliance with Ethical Standards}
This research study was conducted retrospectively using human subject data made available in open access by the Pangloss Collection 
(Vatlongos: \href{https://creativecommons.org/licenses/by-nc-nd/3.0/}{CC BY-NC-ND 3.0}, Nashta: \href{https://creativecommons.org/licenses/by-nc/2.5/}{CC BY-NC 2.5}, Kakabe: \href{https://creativecommons.org/licenses/by-nc-nd/3.0/}{CC BY-NC-ND 3.0}), Tokyo University of Foreign Studies
(\href{https://creativecommons.org/licenses/by-sa/4.0/deed.en}{CC BY-SA 4.0}),
and Panayotov et al.~\cite{panayotov2015librispeech} (\href{https://creativecommons.org/licenses/by/4.0/}{CC BY 4.0}). 
Ethical approval was not required as confirmed by the license attached with the open access data.


\bibliographystyle{IEEEbib}
\bibliography{strings,refs}

\end{document}